\pdfoutput=1

\documentclass[11pt]{article}

\usepackage[]{acl2022}

\usepackage{times}
\usepackage{latexsym}

\usepackage{graphicx}

\usepackage{booktabs}
\usepackage{multirow}

\usepackage[T1]{fontenc}

\usepackage[utf8]{inputenc}

\usepackage{microtype}

\newcommand{\psa}{\textsc{psa}}
\newcommand{\ps}{\textsc{ps}}
\newcommand{\p}{\textsc{p}}

\title{Single-Turn Debate Does Not Help Humans Answer Hard Reading-Comprehension Questions}

\author{Alicia Parrish,$^1$* Harsh Trivedi,$^2$* Ethan Perez,$^1$* Angelica Chen,$^1$ \\\bf Nikita Nangia,$^1$ Jason Phang,$^1$ Samuel R. Bowman$^1$  \AND
\textnormal{$^1$New York University} \And
\textnormal{$^2$Stony Brook University} \AND 
\textnormal{\normalsize Correspondence: {\tt \href{mailto:alicia.v.parrish@nyu.edu}{alicia.v.parrish@nyu.edu}, \href{mailto:bowman@nyu.edu}{bowman@nyu.edu}}}}

\begin{document}
\maketitle

{
\let\thefootnote\relax\footnote{*~Equal contribution.} 
}

\begin{abstract} 

Current QA systems can generate reasonable-sounding yet false answers without explanation or evidence for the generated answer, which is especially problematic when humans cannot readily check the model's answers.
This presents a challenge for building trust in machine learning systems.
We take inspiration from real-world situations where difficult questions are answered by considering opposing sides \cite[see][]{irving2018ai}.
For multiple-choice QA examples, we build a dataset of single arguments for both a correct and incorrect answer option in a debate-style set-up as an initial step in training models to produce \textit{explanations} for two candidate answers. 
We use long contexts---humans familiar with the context write convincing explanations for pre-selected correct and incorrect answers, and we test if those explanations allow humans \textit{who have not read the full context} to more accurately determine the correct answer.
We do not find that explanations in our set-up improve human accuracy, but a baseline condition shows that providing human-selected text snippets does improve accuracy.
We use these findings to suggest ways of improving the debate set up for future data collection efforts.

\end{abstract}

\section{Introduction}

Challenging questions that humans cannot easily determine a correct answer for (e.g., in political debates or courtrooms) often require people to consider opposing viewpoints and weigh multiple pieces of evidence to determine the most appropriate answer.
We take inspiration from this to explore whether debate-style explanations can improve how reliably humans can use NLP or question answering (QA) systems to answer questions they cannot readily determine the ground-truth answer for.

As QA models improve, we have the opportunity to use them to aid humans, but current models do not reliably provide correct answers and, instead, often provide believable yet false responses \cite[][i.a.]{nakano2021webgpt}.
Without access to the ground truth, humans cannot directly determine if an answer is false, especially if that answer comes with a convincing-sounding explanation.
A solution could be for QA systems to generate explanations with evidence alongside different answer options, allowing humans to serve as judges and assess the validity of the model's competing explanations \cite{irving2018ai}.
This approach may be most useful when humans cannot readily determine the ground truth.
This is the case for dense technical text requiring expert knowledge and for long texts where the answer is retrievable, but it would take significant time; we consider the latter as a case study.

We create a dataset of answer explanations to long-context multiple choice questions from QuALITY \cite{pang2021quality} as an initial step in this direction.
The explanations are arguments for pre-determined answer options; crucially, we collect explanations for both a correct and incorrect option, each with supporting evidence from the passage, to create debate-style explanations.
To assess the viability of this data format, we test if humans can more accurately determine the correct answer when provided with debate-style explanations.

We find that the explanations do not improve human accuracy compared to baseline conditions without those explanations.
This negative result may be specific to the chosen task set-up, so we report the results and release the current dataset as a tool for future research on generating and evaluating QA explanations. 
We offer concrete suggestions for future work that builds on the current dataset and alters the task set up in a way that allows humans to more accurately determine the correct answer.
The ultimate goal is to develop a fine-tuning dataset for models that can both explain why a potential answer option is correct and cite the evidence that is the basis for that explanation in a way that humans find understandable and helpful, \textit{even in the context of an unreliable system}. 

\begin{table*}[h!]
    \newcommand{\argwidth}{23.5ex}
    \newcommand{\snippetwidth}{26.5ex}
    \newcommand{\doubletabwidth}{50ex}
    \footnotesize
    \centering
    \begin{tabular}{p{\argwidth}p{\snippetwidth}p{\argwidth}p{\snippetwidth}}
        \toprule
         \multicolumn{4}{c}{\textbf{Question}: \textit{What clearly showed a sense humbleness presented by Si?}} \\
         \multicolumn{2}{p{\doubletabwidth}}{\textbf{Correct option}: \textit{His lack of awareness that he would be considered a celebrity at the Kudos Room.}} & \multicolumn{2}{p{\doubletabwidth}}{\textbf{Incorrect option}: \textit{His quaint behavior at the banquet where he was presented with a gold watch.}} \\
         \midrule
         \textbf{Argument}\newline Si clearly puts the Kudos Room on a pedestal as a place for the top echelons of society and feels humbled to be sitting there, even thinking back to how he dreamed about it while sitting in his space craft (\#1). He seems taken aback when Natalie recognises him as the famous space pilot and even seems to downplay his status and accomplishments (\#2). While Natalie seems star-struck by his presence, he seems equally star-stuck by her beauty, showing how humble he is despite being famous (\#3). & \textbf{Text snippets} \newline (1) Well, this was something like it. This was the sort of thing he'd dreamed about, out there in the great alone, seated in the confining conning tower of his space craft. He sipped at the drink, finding it up to his highest expectations\newline (2) The girl, her voice suddenly subtly changed, said, "Why, isn't that a space pin?" Si, disconcerted by the sudden reversal, said, "Yeah ... sure."\newline (3) Imagine meeting Seymour Pond. Just sitting down next to him at a bar. Just like that. "Si," Si said, gratified. Holy Zoroaster, he'd never seen anything like this rarified pulchritude. Maybe on teevee & \textbf{Argument}\newline It's clear from \#1 and \#2 that in the professional world in which Si moved, a high standard of living was expected. Symbols of prestige were also considered desirable in this social world, reflected by him being awarded a gold watch (see \#3). However, it's clear that Si doesn't care for symbols of prestige like gold watches, prefer more practical items instead Nor is he desirous of a higher standard of living. He only wants enough money to meet life's necessities. & \textbf{Text Snippets} \newline (1) They hadn't figured he had enough shares of Basic to see him through decently. Well, possibly he didn't, given their standards. But Space Pilot Seymour Pond didn't have their standards.\newline (2) He'd had plenty of time to think it over. It was better to retire on a limited crediting, on a confoundedly limited crediting, than to take the two or three more trips in hopes of attaining a higher standard.\newline (3) In common with recipients of gold watches of a score of generations before him, Si Pond would have preferred something a bit more tangible in the way of reward\\
         \bottomrule
    \end{tabular}
    \caption{Example of opposing arguments, with extracted evidence, for two options to a question from QuALITY about a science-fiction story. The full passage for this example is at \href{https://www.gutenberg.org/ebooks/52995}{gutenberg.org/ebooks/52995}.}
    \label{tab:example-args}
\end{table*}

\section{Related Work}

Prior work has explored using models to generate explanations~\cite{camburu2018esnli,rajani-etal-2019-explain,zellers2019from}, but there is limited work on using those explanations to verify the model’s prediction, particularly when a human cannot perform the task directly.
Such a dataset would be useful, as model explanations can aid humans in tasks such as medical diagnosis~\cite{cai2019hello,lundberg2018explainable}, data annotation~\cite{schmidt2019quantifying} and deception detection~\cite{lai2019on}.
However,~\citet{bansal2021does} highlight that these studies use models that outperform humans at the task in question, undermining the motivation for providing a model's explanation alongside its prediction.
When the performance of models and humans is similar, current explanation methods do not significantly help humans perform tasks more accurately \citep{bansal2021does}.
However, explanations based on a mental model of the human's predicted actions and goals can reduce task completion time \citep{gao-2020-joint}. 
We address these shortcomings by collecting data for training models to provide explanations on tasks that would otherwise be time-consuming for humans. 

In addition to task characteristics, several qualities of the model explanation affect the helpfulness of human-AI collaboration: 
Machine-generated explanations only improve human performance when the explanations are not too complex \cite{ai-2020-beneficial,narayanan-2018-how}. 
And though users want explanations of how models mark answers incorrect, most explanations that models output focus on the option selected \cite{liao2020questioning}. 
Our dataset addresses this by including evidence and explanations for both correct and incorrect options to each question, enabling models trained on it to present arguments for more than one answer.


\section{Argument Writing Protocol}

We build a dataset of QA \mbox{(counter-)explanations} by having human writers read a long passage and construct arguments with supporting evidence for one of two answer options.
We then present the explanations side-by-side to a human judge working under a strict time constraint, who selects which answer is correct given the two explanations.

\paragraph{Passage and Question Selection}
We use passages and questions from a draft version of the recent long-document QA dataset, QuALITY \cite{pang2021quality}. 
In QuALITY, most passages are science fiction stories of about 5k words with 20 four-option multiple-choice questions.
We determine which of the three incorrect options is best suited to have a convincing argument by identifying cases where (i) humans in a time-limited setting incorrectly selected that choice at least 3/5 times, and/or (ii) humans who read the entire passage selected that choice as the best distractor item more than half the time. 
We discard questions without an incorrect answer option meeting either criteria. 


\paragraph{Writing Task}
We recruit 14 experienced writers via the freelancing platform Upwork (writer selection details are in Appendix \ref{app:writing-task-details}). 
We assign each writer up to 26 passages. 
Each passage has 7--15 2-option multiple choice questions (avg. of 13.3). 
We have writers construct an argument (max 500 characters) and select 1--3 supporting text snippets (max 250 characters) for one of those two options (Table~\ref{tab:example-args}), with the rate of correct and incorrect options assigned to each writer roughly equal. 

We encourage writing effective arguments by awarding writers a bonus each time a worker in the judging task selects the answer they wrote an argument for. 
Including bonuses, workers average \$21.04/hr, after taking Upwork fees into account.
Further details are in Appendix \ref{app:writing-task-details}, and a description of the writing interface is in Appendix \ref{app:writer-ui}.

\paragraph{Final Dataset}
We release a dataset of both correct and incorrect arguments with selected text snippets and the results of the judgment experiment as a tool for researchers.
These datasets are available at \href{https://github.com/nyu-mll/single_turn_debate}{github.com/nyu-mll/single\_turn\_debate}.
As we use passages from a draft version of QuALITY, we do not release arguments from passages in their non-public test set.
The final dataset that we release contains 2944 arguments (50\% correct) from 112 unique passages, each with an average of 2.4 text snippets. 

\section{Judging Protocol}

We test the effectiveness of the arguments by having human judges answer the multiple-choice question.
To ensure that the judges cannot simply read the passage to find the answer themselves, we give them only 90 seconds of access to the passage along with the arguments and text snippets.
To determine whether the arguments affect human accuracy, we compare the performance of workers who see those arguments and snippets to the performance of workers who do not see the arguments and workers who see neither the arguments nor the text snippets.



\paragraph{Judging Task Protocol}
We recruit 194 workers via Amazon Mechanical Turk (MTurk; recruitment details are in Appendix \ref{app:judge-recruitment}).
Each worker judges which of two answer options is correct, given just 90 seconds.
The worker has unlimited time to read the question and answer options before starting a 90-second timer.
Once the timer is started, the worker can view the entire passage, as well as the arguments and text snippets for each answer option.
Clicking on the snippets scrolls to and highlights the relevant section of the passage so that the snippet can be viewed in context.
Once the timer runs out, the worker has 30 seconds to finalize their answer before the task auto-submits, though workers can submit their answer at any time. 
After submitting, workers see immediate feedback about their accuracy to help them improve over time and to increase engagement.
Each question is judged by three unique workers, and we ensure workers are paying attention with catch trials (Appendix~\ref{app:catch-trials}).
Details on the judging interface are in Appendix~\ref{app:judge-ui}.

\paragraph{Payment and Bonus Structure}
Workers receive \$0.15 per task and a bonus of \$0.40 for each correct answer. 
We aim for the low base pay and generous bonuses to disincentivize guessing.
Assuming workers spend 90 seconds per task, including reading the question and answer options,\footnote{Median completion times after starting the timer were about 60s, so total completion times were likely $<$90s.} a worker with an accuracy of 65\% earns \$16.40/hr.

\paragraph{Baselines}
We include two additional conditions to better understand the effects of arguments in this time-limited setting.
The main protocol is the \textbf{passage+snippet+argument condition (\psa)}. 
The baselines present just the \textbf{passage+snippet (\ps)} or just the \textbf{passage with no supporting evidence (\p)}. 
All other details of the protocol remain the same. 
Each worker only sees tasks in one condition at a time, but through three rounds of data collection, they alternate through the conditions in a random and counterbalanced way.
No worker judges the same question in multiple conditions.

\paragraph{Pilot Judges}
During the writing phase, we use a smaller pool of workers who we qualify as an initial group of judges to gather feedback for the writers and determine their bonuses.
In this group, five judges rate each question, and we test the effects of different time limits, which vary in different rounds between 60, 90, or 120 seconds. 
These pilot results are not part of our main results, but we include the pilot results and details about the pilot judges in Appendix \ref{app:diff-time-limits}.
All other task details are the same as for the main judges.

\section{Results}

In addition to the primary comparison across conditions, we conduct exploratory analyses to better understand effects of the task set-up on workers' response behavior. 
Results on features of arguments and text snippets are in Appendix \ref{app:additional-results}.

\begin{figure}
    \centering
    \includegraphics[width=0.9\linewidth]{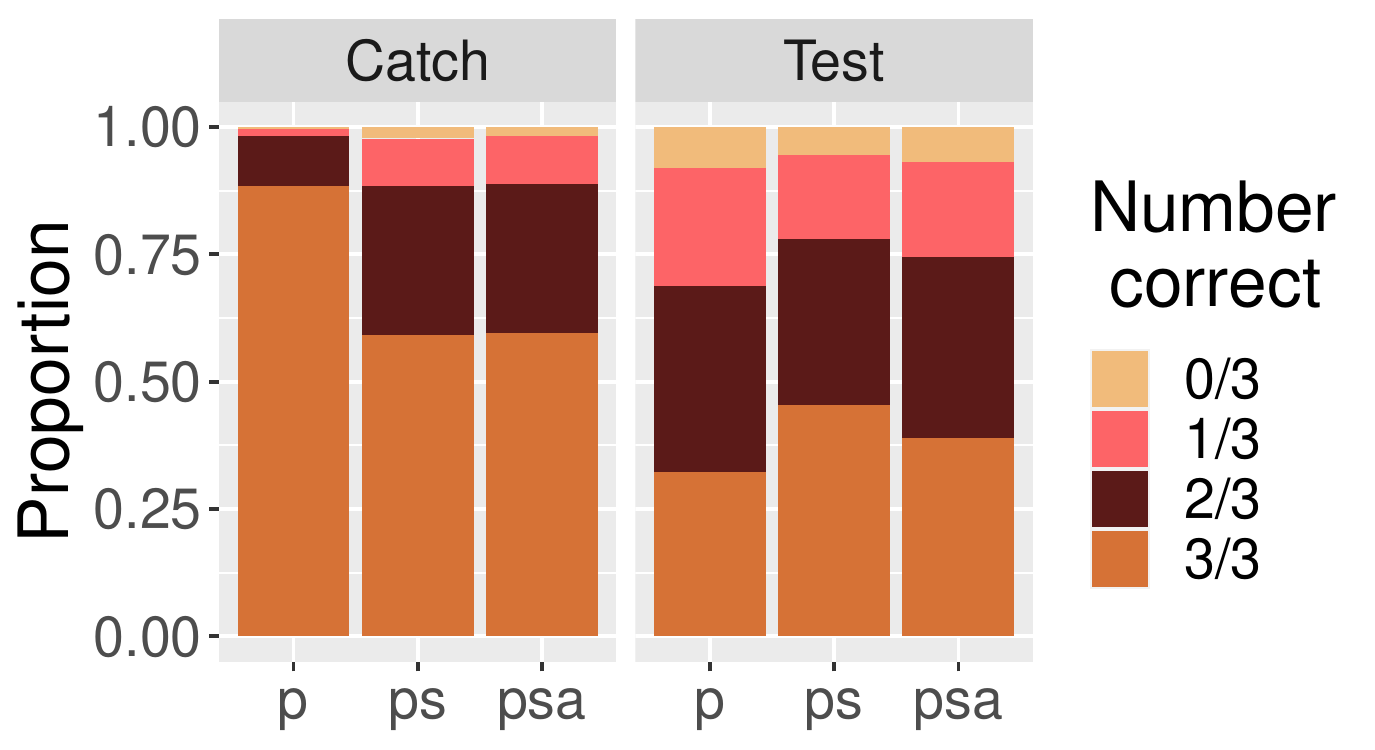}
    \caption{Proportion of workers who answered each question correctly in each condition. \p\ is passage; \textsc{s} is snippets; \textsc{a} is arguments}
    \label{fig:mainresults}
\end{figure}

\paragraph{Comparison Across Conditions}
Workers are more accurate when they have access to text snippets, and they are the most accurate in the \ps\ condition, indicating no clear effect of the arguments.
Figure \ref{fig:mainresults} shows the accuracy rates by question in each of the conditions. 
Both unanimous agreement (3/3 workers correct) and majority vote agreement ($\ge$2/3 workers correct) show that workers are most accurate in \ps\ and least accurate in \p.

\paragraph{Effects of Time}
We investigate if workers get more accurate at this task over time to see if they are learning task-specific strategies. 
Workers' accuracy does improve slightly over time, by about 4 percentage points in each condition between the first 10 tasks and final 10 (Appendix \ref{app:additional-results}, Figure~\ref{fig:improvements-over-time}).
The accuracy increase is small and could be accounted for by workers becoming more familiar with the task format or by figuring out a moderately effective strategy.

Most workers submit an answer before the 90s timer ends. 
Median completion times are longest in \p\ (69s) and similar between \ps\ (54s) and \psa\ (57s).
The average time spent varies by worker, so we check if spending more time leads to higher accuracy. 
However, there is no correlation between workers' average task time and average accuracy (Appendix \ref{app:additional-results}, Figure~\ref{fig:acctime-corr}).

\paragraph{Follow-up Survey}
We release a paid survey to workers who completed at least 10 tasks in each condition to ask about what strategies they used and to better understand their reactions to the arguments.
102 workers qualified for the survey, and 91 completed it.
Workers who reported reading the snippets had significantly higher accuracy in \ps\ and \psa\ compared to workers who did not report reading them. 
However, there are no significant differences in \psa\ accuracy based on whether the workers reported reading the arguments or ignoring them.
A quarter of workers reported mistrusting the arguments; though mistrust does not correlate with performance, see Appendix \ref{app:additional-results} for discussion.


\section{Discussion}
We find it likely that explanations will be beneficial to users in \textit{some} tasks under \textit{some} conditions. 
The prevalence of a debate-style set up in real-world settings (e.g., courtrooms\footnote{We are \textit{not} suggesting this be used in \textit{actual} courtrooms.}) makes this an \textit{a priori} reasonable area for systematic exploration, but the current study is limited in its scope and is not strong evidence against the broad potential usefulness of such a set-up.
The current experiments are a case study in creating a scenario where humans are \textit{unable} to be sure about their answer, but they have access to evidence to help identify the correct response. 
The finding that a quarter of workers mistrusted the arguments raises the issue of whether an approach that gives users misleading information from the outset is on the wrong track.
However, we already know QA models provide false and misleading information; this behavior has the potential to be \textit{more} harmful when it is not explicit that generated explanations may be wrong.  

One reason that the arguments were more misleading than helpful to some workers could be that the correct and incorrect arguments were \textit{independent} of each other.
The strength of debate for determining the true answer could rely on counter-arguments that explicitly reference deficiencies of the other argument.
It is therefore possible that a \textit{multi-turn} setting is needed for debate to be helpful, but we leave this as a question for future research.

The time limit that we use makes the task more artificial than we'd like.
However, pilot results (Appendix \ref{app:diff-time-limits}) show that variations between 60 and 120 seconds make virtually no difference in performance.
It is possible that 120s is still too short, and so workers rushed through the task as much as they did with 60s, but we would have expected this to vary more by worker, and the general trend is that people are slightly \textit{less} accurate at 120s than at 90s.

\section{Conclusion}

We set out to test whether providing users with arguments for opposing answer options in a multiple choice QA task could help humans be more accurate, even when they haven't read the passage.
The results indicate that the task set up had little to no effect on accuracy, but it raises new questions and possible future directions for when such explanations may be useful.

\section*{Acknowledgements}

This project has benefited from financial support to SB by Eric and Wendy Schmidt (made by recommendation of the Schmidt Futures program), Samsung Research (under the project \textit{Improving Deep Learning using Latent Structure}) and Apple. This material is based upon work supported by the National Science Foundation under Grant Nos. 1922658 and 2046556. Any opinions, findings, and conclusions or recommendations expressed in this material are those of the authors and do not necessarily reflect the views of the National Science Foundation. 

\bibliography{anthology,custom}
\bibliographystyle{acl_natbib}

\appendix

\section{Writing Task Details}
\label{app:writing-task-details}
\paragraph{Writer Recruitment}
We list our task on the freelancing platform Upwork as a writing job open to all workers.
We received 112 applications and selected 26 of the most qualified writers to complete a qualification task (2 chose not to complete the qualification).
The 24 writers who finish the qualification task are paid \$36.00 to complete (i) a tutorial task that consists of a full passage and 10 example arguments with supporting text snippets and explanations about how each argument is constructed, followed by (ii) a qualification task that consists of reading a new passage and constructing 10 arguments with supporting text snippets.
Each submission is evaluated on a numeric scale by two of the authors and rated for how convincing the argument is, how useful the snippets are, and how closely the argument needs to be read to select that answer or exclude the other answer option (in order to make sure the writers can construct clear and concise arguments).
We aggregate these results for each writer by $z$-scoring the ratings by each evaluator's scores, and then averaging across questions for each metric.
We select the top-performing 14 writers to continue on to the main writing task.

\paragraph{Pay and Bonus Structure}
We pay writers a base rate of \$18 per passage. 
As it is more difficult to write a convincing explanation for an incorrect answer compared to a correct one, we award writers a bonus of \$0.10 for each time a judge selects their argument for a correct answer and \$0.50 for each time a judge selects their argument for an incorrect answer option.
Which answer option is correct and which one is incorrect is not revealed to the writers during the writing task; they only see this information once they receive feedback about how the judges performed, at which point they find out how much of a bonus they earned.

As stated in the main text, each passage in our final dataset has 7--15 2-option multiple choice questions (avg. of 13.3).
However, in the full task given to writers, they constructed arguments for 11-15 questions per passage (average 14.2), but we later determined from metadata in QuALITY that some questions were ambiguous, and we removed those questions from the dataset. 

Each multiple choice question is judged by 5 different crowdworkers (see Appendix \ref{app:initial-judges} for information on these judges), 
and the average bonus rate per passage is \$7.43 (range \$2.90 - \$15.30), for an effective average hourly rate\footnote{We estimate it takes one hour to complete each passage based on pilot runs and discussion with the writers} of \$21.04/hr after taking into account Upwork fees.\footnote{Unlike other crowdsourcing platforms like MTurk, Upwork charges fees on the worker's end, and these fees change depending on how much has already been paid to that worker.}

\section{Writer Interface}
\label{app:writer-ui}

The interface for writers includes a dashboard where the writer can view the passages that we assign them, along with a progress bar for that batch of work. 
Each passage contains a pane with the full passage and another pane with the questions with both answer options. 
Writers select text snippets by highlighting the relevant portion of the passage and clicking an 'add snippet' button.
Writers are restricted from writing arguments longer than 500 characters or text snippets longer than 250 characters to encourage conciseness and to ensure that judges will be able to read the arguments within the time limit. 
The writer must both write an argument and select at least one text snippet for each answer.
In order to keep the method of referencing text snippets as consistent as possible across different writers with the ultimate goal of being able to train an LM to generate similar arguments, we instruct the writers that they should reference the snippets they select in a uniform way, by either referring to the argument as `\#1' or by placing the argument number in parentheses after the relevant part of the argument, as if it were a citation.

\begin{figure*}
    \centering
    \includegraphics[width=0.95\textwidth]{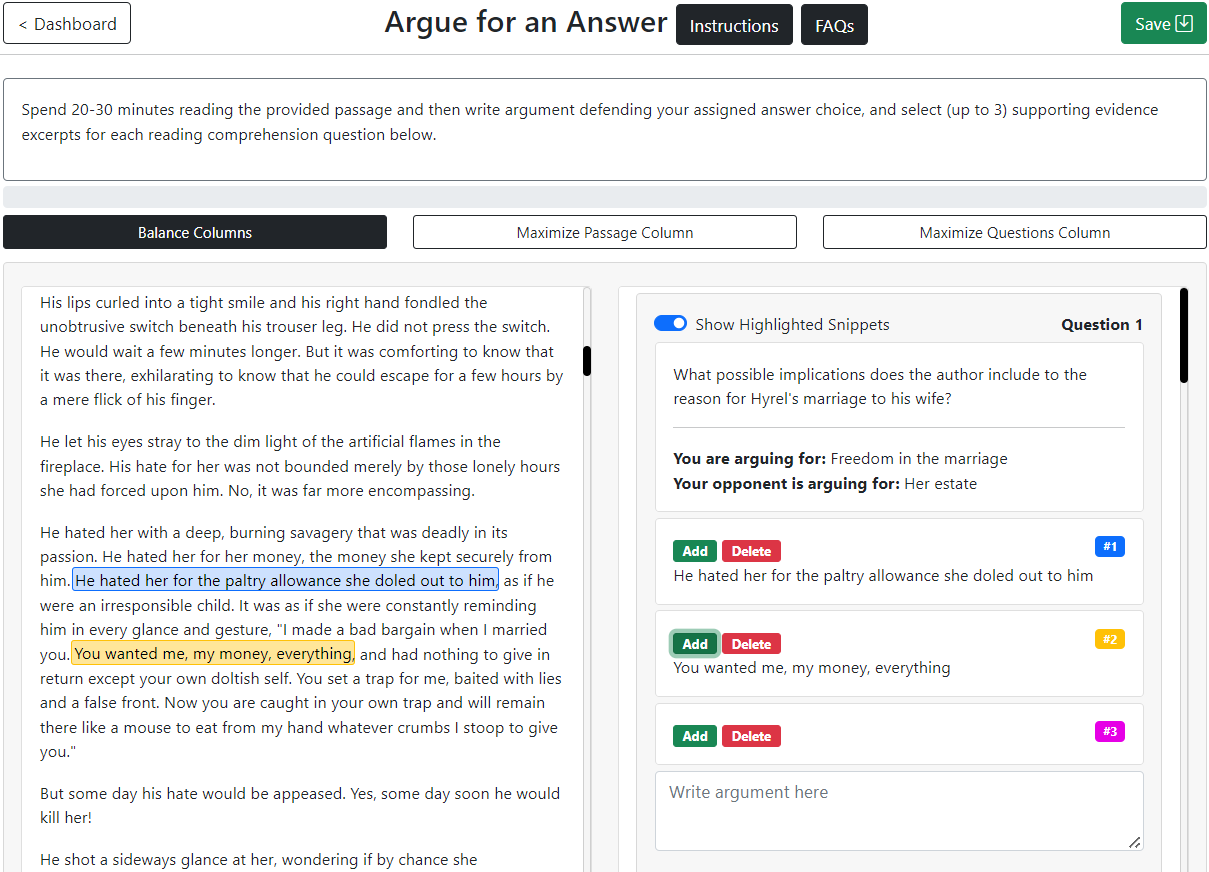}
    \caption{Argument writing interface. In this example, two text snippets have been selected for Question 1.}
    \label{fig:writer-ui}
\end{figure*}

Once all the arguments have gone through the judging phase, the writers can view the feedback via their dashboard to see how each of their arguments performed.
This dashboard lists how many judges from the \psa\ condition chose their argument, along with how much of a bonus they earned.
This feedback remains available to the writers as they write the next round of arguments.

\section{Judging Task Crowdworker Recruitment}
\label{app:judge-recruitment}
We recruit judges via Amazon Mechanical Turk (MTurk) using a question-answering qualification task that is open to workers with at least a 98\% HIT approval rating and at least 5000 HITs completed; this task pays \$2, with a bonus of \$1 for anyone who passes, and takes approximately 8-10 minutes to complete.
In this task, workers read 5 passages of 105--184 words and then answer 2 four-option multiple choice questions about each.
A total of 400 workers complete this task, and 249 of them achieve an accuracy above the threshold of 90\%. 
Of these qualified workers, 194 of them end up completing the main task.

\section{Judging Interface}
\label{app:judge-ui}

\begin{figure*}
    \centering
    \includegraphics[width=0.95\textwidth]{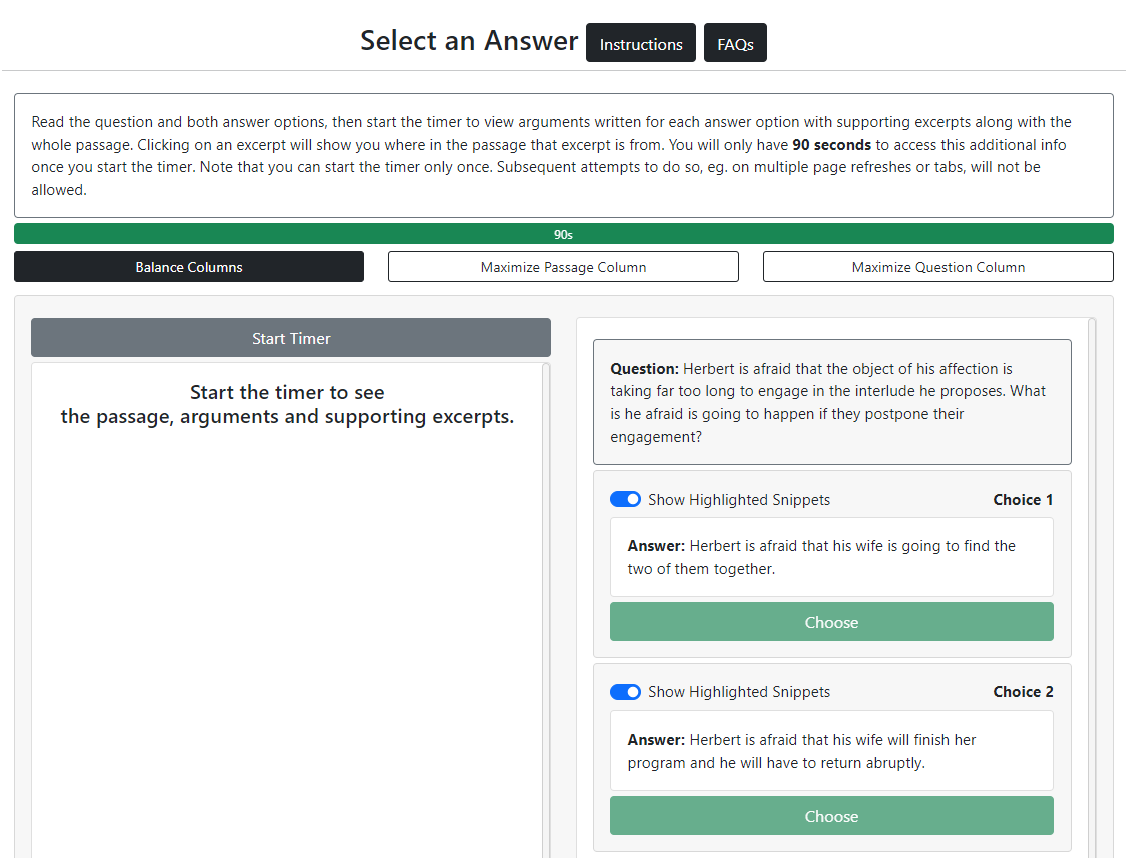}
    \caption{Judging UI before starting the 90s timer.}
    \label{fig:pre-timer-start-UI}
\end{figure*}

Judging interfaces are mostly the same in each condition, and only vary in what information is revealed when a worker hits the 'start timer' button (in addition to corresponding changes in the instructions).
Figure \ref{fig:pre-timer-start-UI} shows the state of the UI before a worker starts the timer.
At this point, the worker only has access to the question and the two answer options.
The worker is unable to select either option before starting the timer.

\begin{figure*}
    \centering
    \includegraphics[width=0.95\textwidth]{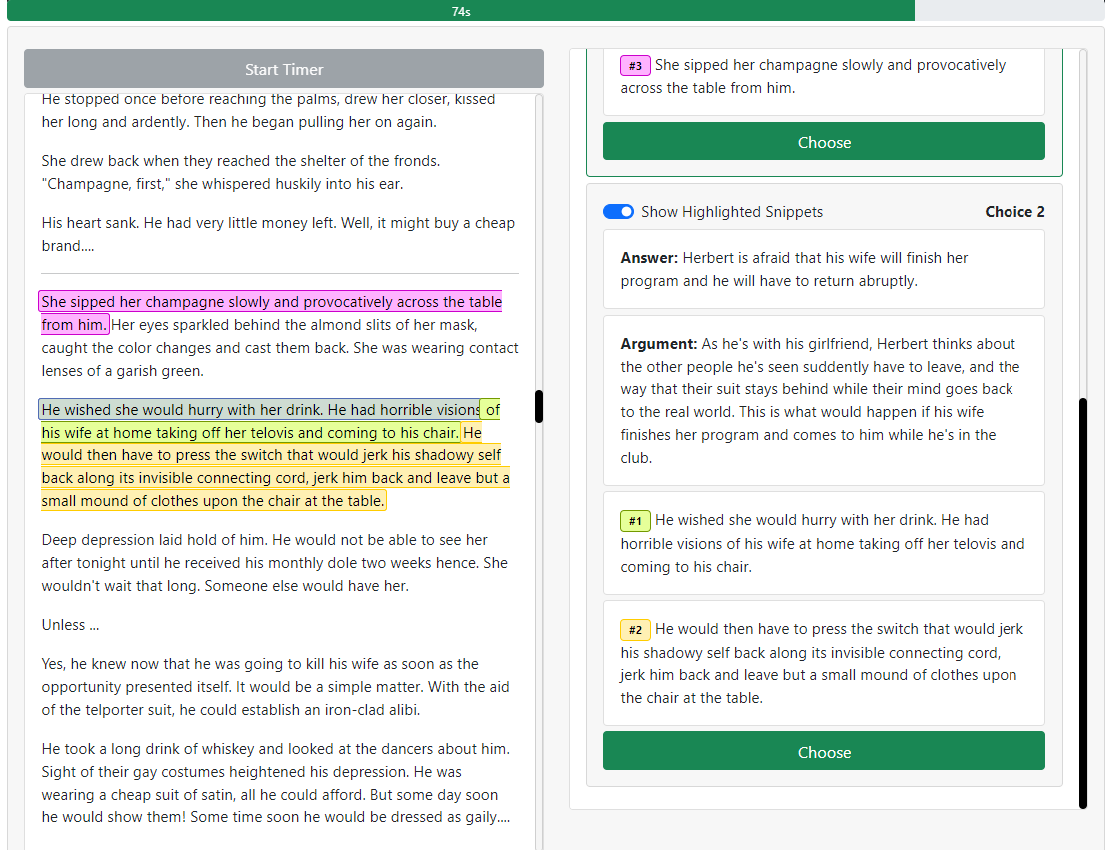}
    \caption{Judging UI after starting the 90s timer. This view shows what happens after someone clicks on one of the text snippets for argument 2 and gets taken to the relevant portion of the text, with that part of the text highlighted.}
    \label{fig:post-timer-start-UI}
\end{figure*}

Figure \ref{fig:post-timer-start-UI} shows an example from \psa\ where after clicking 'start timer,' the passage, text snippets, and arguments for each of the two answer options is revealed.
As the worker scrolls down, the timer remains visible at the top of the screen.
Clicking on any of the text snippets auto-scrolls to the relevant portion of the passage and shows color-coded highlights from the text that match the text snippets under each argument.
After selecting an answer, the worker scrolls to the bottom of the screen to hit the 'submit' button.

\begin{figure*}
    \centering
    \includegraphics[width=0.95\textwidth]{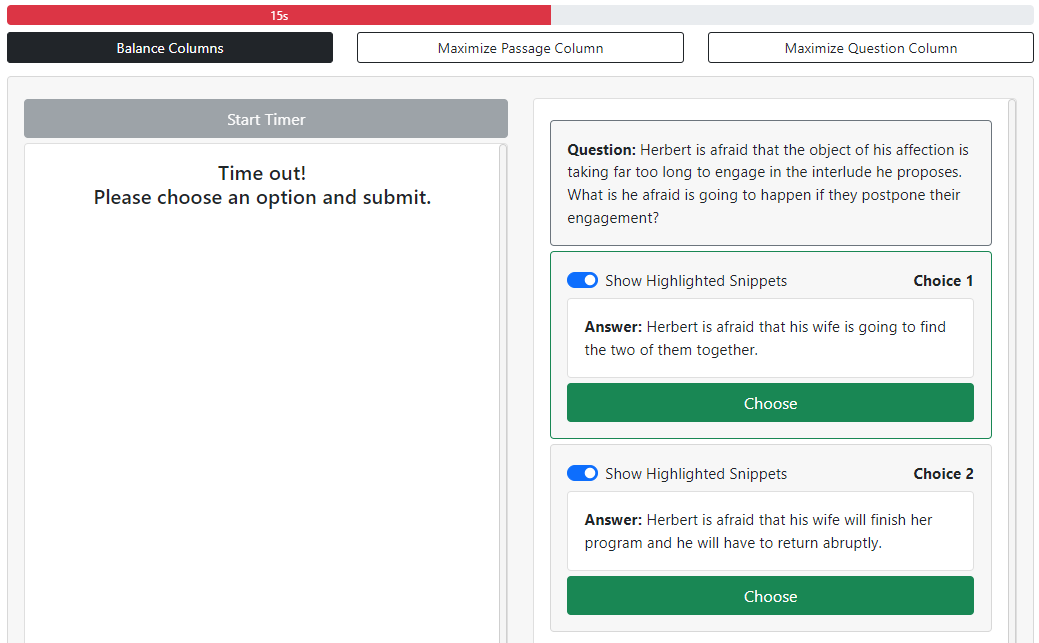}
    \caption{Judging UI after the 90s timer has run out. The arguments, snippets, and text have disappeared, and the judge has only 30 seconds to select a final answer.}
    \label{fig:timer-stop-UI}
\end{figure*}

If the timer runs out and the worker still has not hit the 'submit' button, all the information that was presented when they hit 'start timer' disappears and the worker has 30 additional seconds to select one of the two options and click 'submit,' as shown in Figure \ref{fig:timer-stop-UI}.
If this final timer runs out, the task auto-submits and the response is recorded as having no selection, which we mark as an incorrect response.

\section{Catch Trials}
\label{app:catch-trials}
We use catch trials, tasks that look like the test trials but are specifically constructed to be able to be correctly answered given a short time limit, to assess if workers are paying attention and making an effort in the task.  
In the \p\ condition, the catch trials are taken from the ones used in QuALITY that were constructed to be answerable within one minute by skimming the passage or using a search function (e.g., they include a direct quote that can be searched for with an in-browser search function like ctrl+F).
In the \ps\ and \psa\ conditions, we construct catch trials by mismatching the argument and/or snippet from another question in that passage onto the incorrect answer option.
In this way, it should be obvious to any worker making a faithful attempt at the task which answer option is correct, as one of them is paired with an unrelated argument and/or set of text snippets.

Throughout data collection, we mix approximately 10\% of the tasks with catch trials.
In order to determine which workers maintain the qualification to complete more tasks, we continuously monitor accuracy on these catch trials.
Once workers have completed at least five catch trials in a given condition, if their accuracy on these falls below 60\%, we prevent them from completing any more tasks.
Although this method relies on workers having already completed a significant number of tasks before we have enough data to dynamically restrict them, this does not seem to be a major concern in data quality because (i) very few workers (6.2\%) end up losing the qualification for the task because of low catch trial accuracy, and (ii) aggregation metrics minimize the effect of a few workers not completing the task felicitously.
Among workers who completed at least five catch trials in a given condition, median accuracy on the catch trials is 88.9\%, indicating that the catch trials can generally be answered given the strict time limit, and that most participants consistently put an honest effort towards the task.

\section{Initial Group of Judges}
\label{app:initial-judges}

During the writing rounds, we use a smaller set of workers as judges and collect five annotations per example.
The responses from these judges are used to calculate the writers' bonuses, and this set-up allows us to test out different time limits.

\paragraph{Crowdworker Recruitment}
We recruit judges via MTurk in two phases.
First, we release a reading-comprehension-based qualification task open to workers with at least a 98\% HIT approval rating and at least 5000 HITs completed; this task pays \$5, with a \$3 bonus for passing the qualification.
In this task, workers read a 3500 word passage and then answer 15 four-option multiple choice questions about that passage.
A total of 140 workers completed this task, and 77 of them achieved an accuracy above the threshold of 85\%.

For the second phase of the qualification, workers complete a timed judging tasks with an up-sampled number of catch trials. 
Sixty-eight of the qualified workers completed at least 24 HITs in this second qualification and were considered for inclusion in the main protocol.
In order to pass this second qualification, workers need to achieve above chance accuracy on the test trials in at least two of the three protocols, and they need to answer no more than one catch trial incorrectly.
Based on these cutoffs, we qualify 57 crowdworkers to move on to the main judging task, and we pay them an additional \$3 bonus.
A total of 55 of these workers chose to then take part in the main task, and 42 completed tasks in all three rounds of data collection.

\paragraph{Results with Different Time Limits}
\label{app:diff-time-limits}

\begin{figure}
    \centering
    \includegraphics[width=\linewidth]{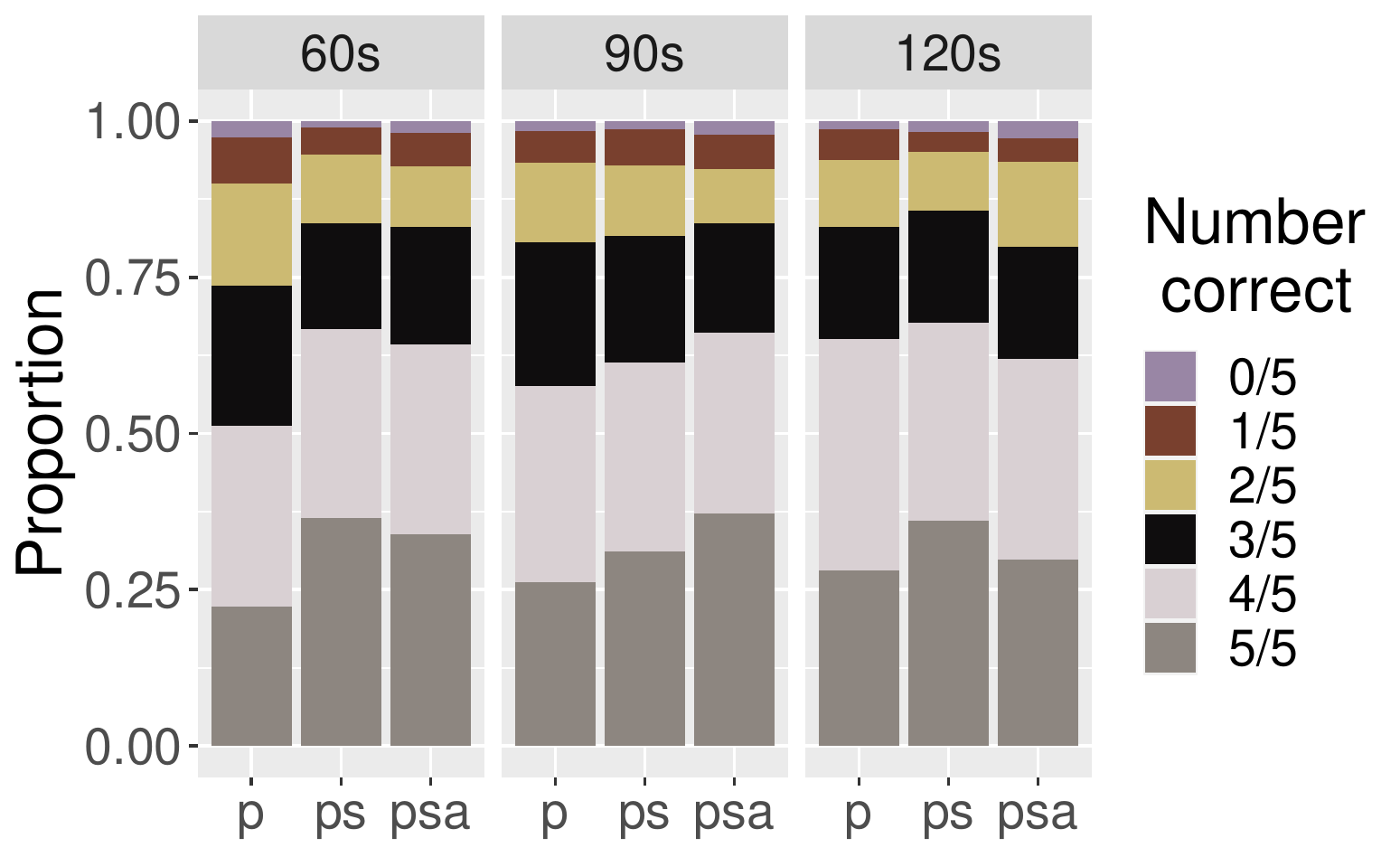}
    \caption{Proportion of pilot judges who answered the question correctly for items within different time limits.}
    \label{fig:by-time-limit}
\end{figure}

During the first round of data collection, we use a 60-second time limit, but we raise this limit to 90 seconds for half of the examples in the second round after feedback from workers indicated that several people in the \psa\ condition did not feel they had sufficient time to read the arguments.
This change resulted in only a very small accuracy increase (see Figure \ref{fig:by-time-limit}), so in the third round, we further raise the time limit for half of the questions to 120 seconds, and keep the 90-second limit for the other half of the questions.
However, the accuracy increase with longer time limits is most pronounced in \p, and so we conclude that performance in \psa\ in particular is likely not strongly driven by how much time workers have to read the arguments.

\section{Effect of Question Selection Method}

As the incorrect answer option was selected based on whether that option was a good distractor in the time-limited validation used by \citet{pang2021quality} \textit{or} based on whether validators who had read the entire passage found that option to be the best distractor, we examine the effect of these two different ways of selecting the incorrect answer option.
In about half of the examples, the incorrect option matched both of these criteria.
Table \ref{tab:distractor-selection} shows that workers are slightly less accurate on questions that were selected as the best distractor by the untimed validators (the ones who had read the entire passage). 
As this difference in accuracy is present in all three conditions and is not more pronounced in \psa\ compared to the other conditions, it is unlikely that this difference is due to the writers being able to construct a better argument for these questions. 

\begin{table}[]
    \centering
    \begin{tabular}{llr}
    \toprule
         Condition & Incorrect & Accuracy \\
         {} & selection & (\%)\\
         \midrule
         \p & both & 68.0\\
         \p & time-limited only & 70.2 \\
         \p & untimed only & 62.5\\
         \ps & both & 73.3\\
         \ps & time-limited only & 74.0 \\
         \ps & untimed only & 72.3 \\
         \psa & both & 71.7 \\
         \psa & time-limited only & 71.2 \\
         \psa & untimed only & 67.7 \\
         \bottomrule
    \end{tabular}
    \caption{Accuracy split by the way the incorrect answer option was selected from among three possible options.}
    \label{tab:distractor-selection}
\end{table}

It's worth noting that we would expect the opposite effect of what we observe for \p, as this condition is identical to the time-limited task used by \citet{pang2021quality}, with the caveat that they showed workers four answer options and those workers had even less time to search the passage.
We do not have a compelling explanation for this result, though it may be that having given workers more time and fewer options to select from allowed them to more accurately identify the answer in these cases because they had more time to search for the answer and had two fewer answer options, which reduced the number of words to use as search terms and made the task substantially easier.
However, this explanation does not account for why accuracy on the questions selected based on QuALITY's time-limited task is the \textit{highest}.

\section{Per-Worker Results}

We observe a great deal of individual variation among workers.
It is likely that some people are better at figuring out what words they need to search for to determine the answer, and there is likely variation in how much workers were able to pick up on patterns that would help them answer correctly.
This variation seems tied to individual variation more than noise from easier vs. harder questions, as we find that an individual's performance in each condition is significantly predictive of their performance in the other conditions, indicating the workers who did well in, for example, \p, were also likely to do well in \ps\ and \psa\ (\p -\ps: \textit{r} = 0.3; \p -\psa: \textit{r} = 0.43; \ps -\psa: \textit{r} = 0.15).

\begin{figure}
    \centering
    \includegraphics[width=0.98\linewidth]{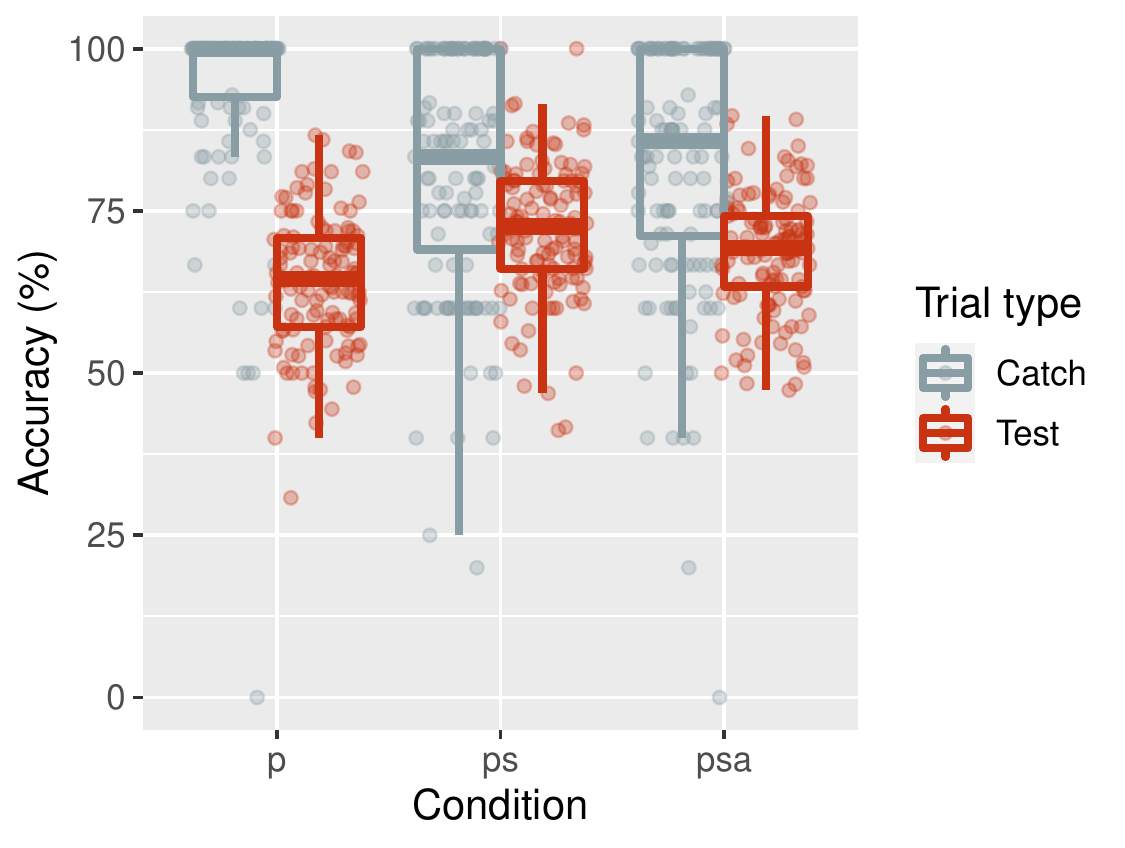}
    \caption{Accuracy of each worker who completed at least 10 tasks in each of the three conditions.}
    \label{fig:by-participant}
\end{figure}




\section{Additional Results}
\label{app:additional-results}

\paragraph{Improvements Over Time}

\begin{figure}
    \centering
    \includegraphics[width=0.98\linewidth]{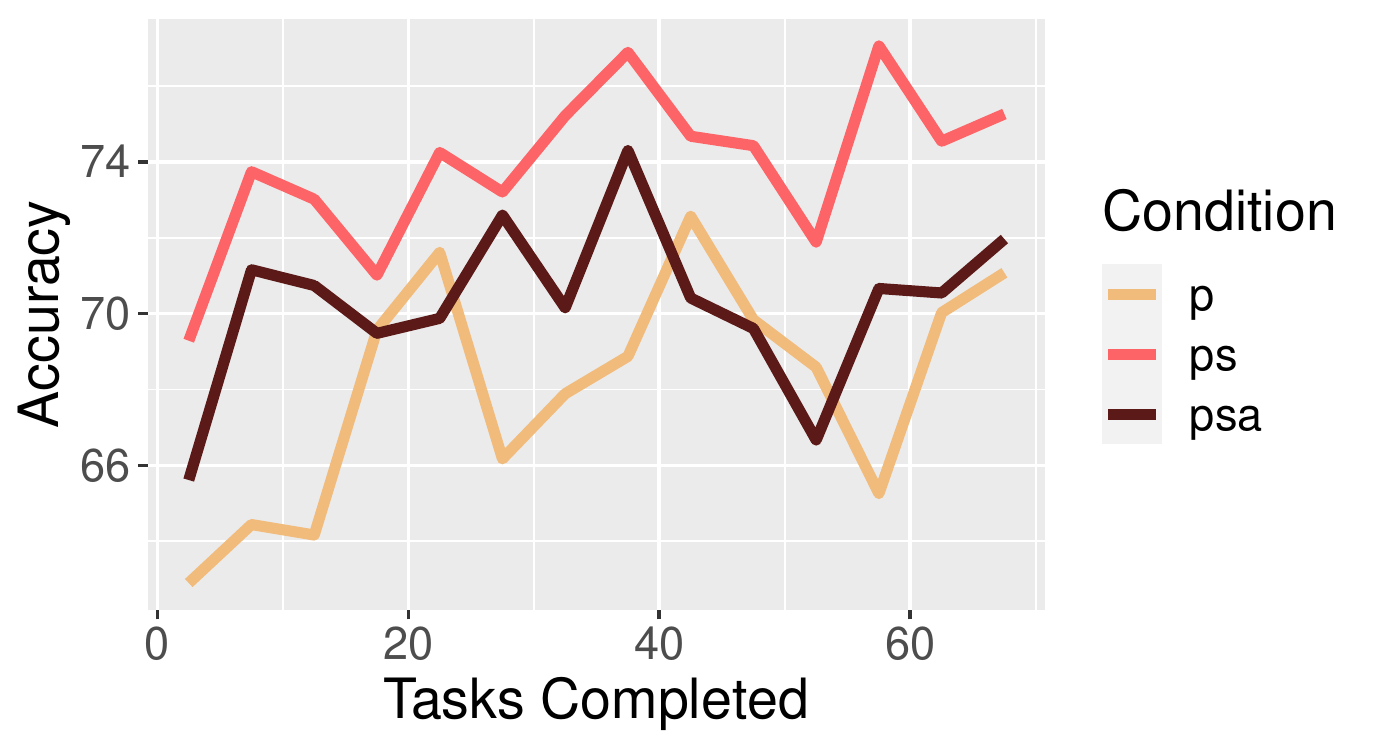}
    \caption{Binned accuracy within each condition, sorted by the order in which each worker completed the tasks. Accuracy improves slightly over time within each condition.}
    \label{fig:improvements-over-time}
\end{figure}

Figure \ref{fig:improvements-over-time} shows the workers' accuracy as they complete more tasks within each condition.
We analyze results for workers who did at least 50 tasks in a given condition.
As workers get more familiar with each condition, their accuracy improves by a total of about four percentage points.
The effect is similar across conditions, and most of the accuracy gains occur after the first 20 tasks completed.

\paragraph{Accuracy by Time Spent on Tasks}

\begin{figure}
    \centering
    \includegraphics[width=0.98\linewidth]{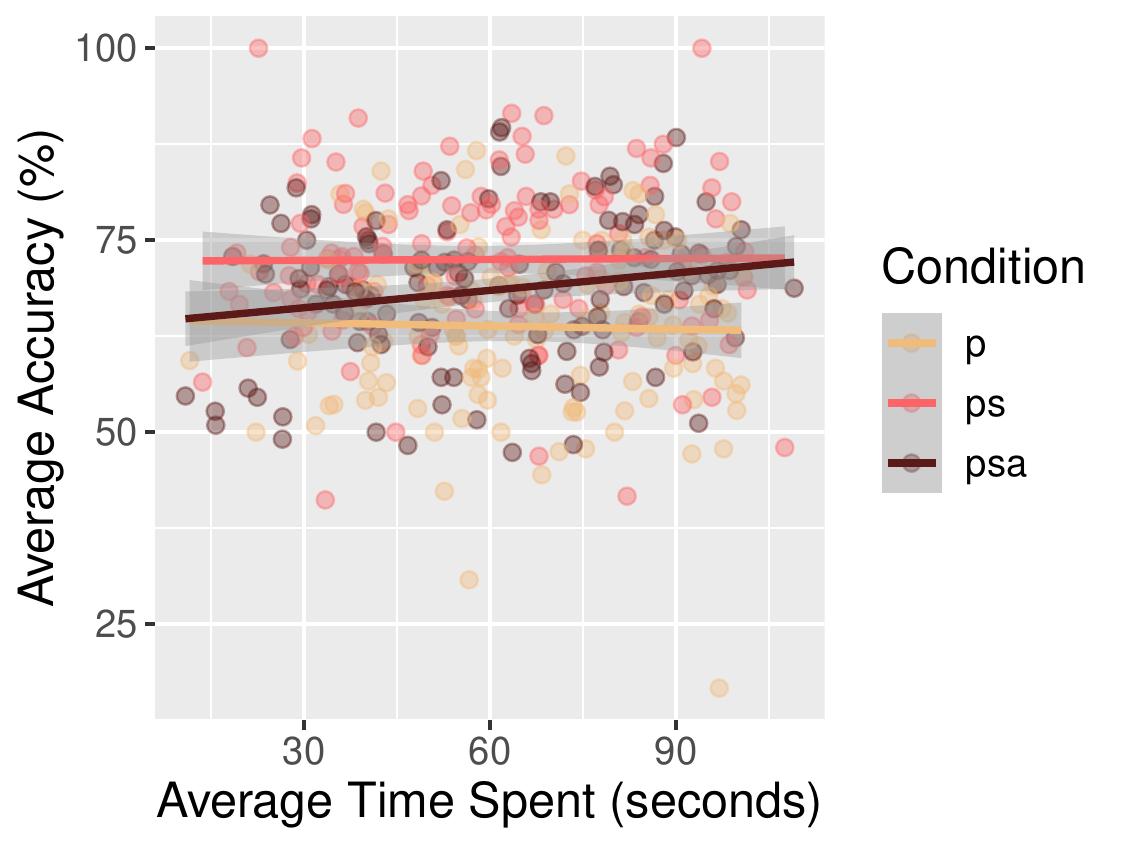}
    \caption{Each worker's average accuracy in each condition, plotted by the average time they spent on each task in that condition. There is no clear advantage to spending more time on the task}
    \label{fig:acctime-corr}
\end{figure}

Figure \ref{fig:acctime-corr} shows the relationship between how long each worker spent, on average, completing each task and how accurate the worker was.
Though there is a very slight positive correlation between time spent and accuracy in \psa, the effect is not statistically significant. 

\paragraph{Length of Arguments and Snippets}
Workers are slightly more likely to choose a longer argument. 
We fit a linear model to predict the rate at which workers choose an answer option from the length of the argument associated with that option in each condition. 
The effect is small, only about a 1.2 percentage point increase in the rate of choosing that option for every 10 additional words in the argument in \psa\ relative to the rate of choosing the same option in \p, but the effect is significant (\textit{p} = 0.001).\footnote{There's no significant difference in argument length based on whether it's arguing for a correct or incorrect answer option.}
Workers are also more likely to choose an answer option supported by more snippets. 
For each additional snippet, there is an increase of 4.2 percentage points in the rate at which workers in \psa\ choose that option, and an increase of 2.8 points in \ps\ (both effects are significantly different from the analogous answer selection rates in \p, \textit{p} < 0.001 and \textit{p} = 0.01, respectively).
 
\paragraph{Effective Argument Words}
We check the most common unigrams within correct arguments, and we find no difference between arguments that were chosen 0, 1, 2, or 3 times by the judges. 
In each case, the four most common words are from within the following set of five words: \textit{earth, time, people, ship, planet}.\footnote{The majority of the context passages were science fiction stories, so these words are expected to come up quite often, relative to their use in other contexts.}
Similarly, the most common bigrams are not frequent enough to be informative, and are often phrases like \textit{time travel} or \textit{main character}.
We also calculate the pointwise mutual information (PMI) of each word within correct and incorrect arguments and within effective and ineffective arguments in order to determine if there are likely to be any lexical regularities workers can pick up on, but no clear trend emerges, and there are numerous ties for words with the highest PMI in each group, even after applying a frequency threshold.

\paragraph{Survey Results Discussion: Mistrust}
Workers are fairly split in whether they found the arguments helpful or generally mistrusted them. 
Though the responses in this survey about the arguments are not predictive of accuracy in any of the three conditions, the responses are useful for considering the more psychological effects of presenting people with arguments we know to be false. 
Having been misled by a convincing-sounding explanation could cause workers to second guess their intuitions and to only rely on information that is grounded in the passage (i.e., the text snippets).
In the survey, nearly a quarter of workers explicitly report mistrusting \textit{and then choosing to ignore} the arguments (51 report choosing to use them, 21 say they either chose not to use the arguments from the beginning or changed tactics halfway through after finding the arguments too misleading, and 19 give responses that can't be coded as either generally trustful/mistrustful).
Although adopting a stance of general mistrust for the arguments is a logical (and perhaps desirable) strategy, the subsequent decision to ignore the arguments entirely due to this mistrust was an unintended consequence of our design.

\end{document}